\documentclass[preprint, 3p, twocolumn]{elsarticle}
\usepackage{graphicx}
\graphicspath{ {images} }
\usepackage{booktabs}
\usepackage{tabularx}
\usepackage[normalem]{ulem}
 \useunder{\uline}{\ul}{}
 \usepackage{rotating}
\usepackage{color}
\def\Manu#1{\textcolor{black}{#1}}

\usepackage[switch]{lineno}
\usepackage{graphicx,xcolor}
\usepackage{epstopdf}
\usepackage{setspace}
\usepackage[T1]{fontenc}
\usepackage{amssymb,amsfonts,amsmath}
\usepackage{setspace}
\usepackage{subcaption}
\usepackage{mathtools}

\begin{document}

\title{\Manu{Towards Real-Time Inference of Thin Liquid Film
Thickness Profiles from Interference Patterns Using Vision Transformers}}
\author[gautam,equal]{G. Anand Viruthagiri}
\author[arnuv,equal]{A. Tandon}
\author[vinny]{G. Fuller}
\author[vinny,vinnyharvard,equal]{V. Chandran Suja\corref{cor1}}\ead{vinny@seas.harvard.edu}
\address[gautam]{John Marshall High School, Rochester, Minnesota 55901}
\address[arnuv]{Department of Computer Science, Stanford University, Stanford, California 94305}
\address[vinny]{Department of Chemical Engineering, Stanford University, Stanford, California 94305}
\address[vinnyharvard]{School of Engineering and Applied Sciences, Harvard University, Boston, Massachusetts 02134}
\address[equal]{Equal contribution}
\cortext[cor1]{Corresponding author}

\begin{abstract}

Thin film interferometry is a powerful technique for non-invasively measuring liquid film thickness with applications in ophthalmology, but its clinical translation is hindered by the challenges in reconstructing thickness profiles from interference patterns - an ill-posed inverse problem complicated by phase periodicity, imaging noise and ambient artifacts. Traditional reconstruction methods are either computationally intensive, sensitive to noise, or require manual expert analysis, which is impractical for real-time diagnostics. To address this challenge, here we present a vision transformer-based approach for real-time inference of thin liquid film thickness profiles directly from isolated interferograms. Trained on a hybrid dataset combining physiologically-relevant synthetic and experimental tear film data, our model leverages long-range spatial correlations to resolve phase ambiguities and reconstruct temporally coherent thickness profiles in a single forward pass from dynamic interferograms acquired \textit{in vivo} and \textit{ex vivo}. The network demonstrates state-of-the-art performance on noisy, rapidly-evolving films with motion artifacts, overcoming limitations of conventional phase-unwrapping and iterative fitting methods. Our data-driven approach enables automated, consistent thickness reconstruction at real-time speeds on consumer hardware, opening new possibilities for continuous monitoring of pre-lens ocular tear films and non-invasive diagnosis of conditions such as the dry eye disease.
\end{abstract}
\maketitle


\section{Introduction}
Thin film interferometry (TFI) is an optical technique widely used to measure film thickness by leveraging interference of light reflected from a film’s top and bottom surfaces, producing characteristic patterns that encode thickness information \cite{born1999principles}. The technique is notable for its ability to non-invasively map thin film topologies with high spatiotemporal resolution \cite{sharma2017idiom, chandran2020hyperspectral, suja2020single}, and is commonly used in applications for studying dynamic processes such as evaporation, spreading, drainage, and rupture in thin film liquid layers \cite{chandran2020hyperspectral, chandran2018evaporation, frostad2016dynamic, vermant2015lung}. In ophthalmology, thin-film interferometry has the potential to be used as an alternative to invasive, low resolution tests, such as the Schirmer’s Tear Test and the fluorescein dye test  \cite{doane1989instrument, bhamla2016instability}, for continuously monitoring tear film stability and diagnosing conditions such as dry eye disease.

Translating this technique to clinical settings is currently hampered by the difficulty in rapidly and accurately reconstructing film thickness from inteferograms, as this process requires solving solving a transcendental phase periodic governing equation \cite{ghiglia1998book, chandran2020hyperspectral}. The phase periodicity leads to a many-to-one mapping between the film thicknesses and interferometric color signatures, making thickness reconstruction from interferograms an ill-posed inverse problem. This coupled with presence of unavoidable imaging noise and additional artifacts, including motion-blur and blinking, makes the thickness reconstruction process ambiguous and challenging for dynamic liquid films. As a result, currently, arbitrary interferograms are analyzed predominantly by a procedure where expert operators track the evolution of fringe patterns over time and space, and utilize this contextual information to resolve the ambiguity that arises from periodicity \cite{frostad2016dynamic, chandran2020hyperspectral}. Though effective, this manual color-matching process is extremely time-consuming, requires extensive expertise, and is subject to inter-operator variability, making this method impractical for high-throughput applications or real-time clinical diagnostics. Conventional techniques attempting to address this bottleneck, such as phase-shifting or iterative model-based fitting, are often computationally intensive, sensitive to noise, or are ill-equipped for analyzing rapidly changing films.
 Phase-shifting methods require multiple images with precisely controlled phase shifts, making them unsuitable for rapidly evolving films and films with an unknown initial thickness. Iterative fitting approaches on the other hand become trapped in local minima due to the phase periodicity, particularly for microscopic films and when imaging noise is present  \cite{chandran2020hyperspectral}. Although implementations of TFI, such as Interferometry Digital Imaging Optical Microscopy (IDIOM), have advanced the visualization of nanoscopic films \cite{ochoa2021foam, sharma2017idiom}, existing methods are broadly inadequate for unconditionally recovering film thickness of arbitrary micron-scale films such as the tear film in real-time \cite{xu2024spatiotemporal,chandran2020hyperspectral}. 

Machine learning (ML) offers a powerful data-driven alternative, bypassing the ill-posed inverse problem by learning the direct, non-linear mapping from raw interferograms to thickness profiles. Similar to how manual reconstruction leverages long-range structure of interferometric fringes and spatial continuity to disambiguate phase periodicity, ML models can learn rich representations that exploit long-range spatial correlations in images to break the ambiguity introduced by noise, imaging artifacts, and phase periodicity. However, unlike manual methods, ML models can perform this pattern recognition automatically, consistently, and at speeds compatible with real-time applications. Early ML approaches in interferometry demonstrated success in relatively simple scenarios, such as using artificial neural networks for thin film thickness measurements to avoid local minima in nonlinear fitting \cite{tabet2000ann, kim2020improved}, and thin-film neural networks (TFNNs) that incorporated transfer matrices as learnable parameters \cite{liu2021tfnn}. Deep learning has proven particularly effective for the challenging problem of phase reconstruction and unwrapping, where neural networks must resolve $2\pi$ ambiguities by recognizing complex spatial patterns \cite{zhong2018machine, signoroni2019deep, wang2019phase, gourdain2022phase}. Comprehensive comparisons have shown deep learning methods significantly outperform traditional algorithms in noisy, discontinuous, and aliasing cases \cite{wang2022comparison}, with temporal phase unwrapping networks demonstrating robustness to intensity noise, low fringe modulation, and motion artifacts \cite{qiao2019temporal}. However, these existing approaches have been primarily developed for static or slowly-varying systems with minimal imaging noise. Achieving accurate profile reconstructions for practical applications remains an open problem, particularly for films with rapidly evolving dynamic processes such as fast evaporation on millisecond timescales, motion artifacts from involuntary blinking, and substantial experimental noise common in clinical data.

To address the above challenges, We introduce a new paradigm for universal interferometry leveraging vision transformers. Our model implementation, trained on a hybrid dataset of synthetic and experimental tear film data, enables the reconstruction of accurate, continuous thickness profiles from noisy, single-shot interferograms in a single, rapid forward pass. The rest of the paper is structured as follows: Section \ref{sec:Methods} details our data generation strategy and the specific adaptations to the network architecture; Section \ref{sec:Results} presents our qualitative and quantitative results demonstrating the performance of the model relative to the state-of-the-art; and Section \ref{sec:Conclusion} concludes with a discussion of the method's impact on clinical tear film analysis.

\begin{figure*}[!h]
\includegraphics[width=\textwidth]{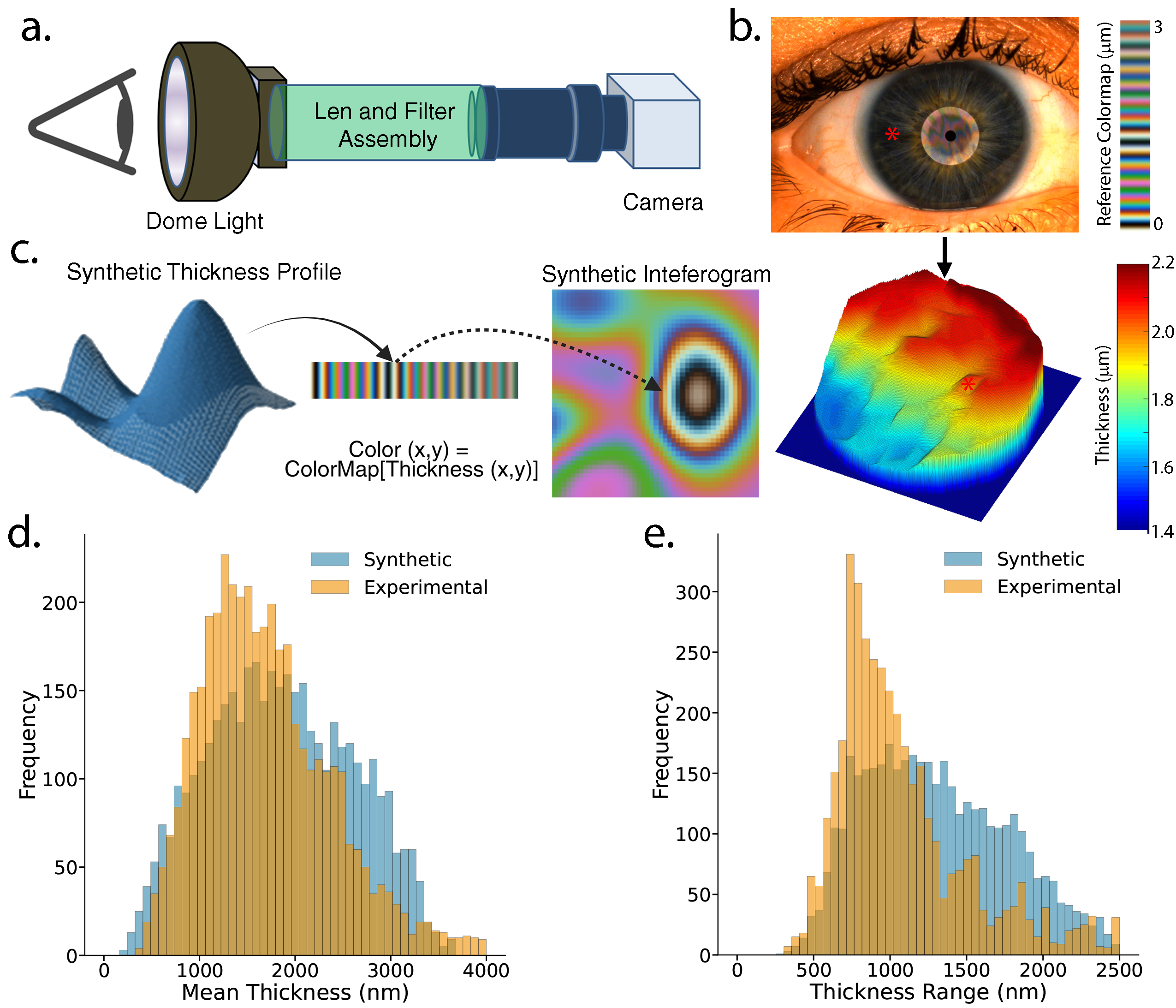}
\caption{Experimental setup and data generation. ({\bf a}) Schematic of the Tear Film Interferometer showing dome light illumination, lens assembly, and camera. ({\bf b}) \textit{In vivo} eye interferogram with reconstructed tear film thickness profile. ({\bf c}) Synthetic data generation workflow: synthetically generated thickness profiles are mapped to color interferograms using a reference colormap to generate training interferograms-thickness profile pairs. ({\bf d}) Mean thickness distribution in the training set. ({\bf e}) Thickness range distribution in the training set. }
\label{fig:InvivoTearFilm_ExperimentalSetup}
\end{figure*}

\section{Methods}\label{sec:Methods}
\subsection{Tear Film Interferometer Setup}

The thin film interferometer arragement developed to image the human tear film is shown in Fig.~\ref{fig:InvivoTearFilm_ExperimentalSetup}a. The system consists of a diffuse dome light (CCS America) providing uniform illumination to generate interference fringes on the eye surface, a precision zoom lens (Edmund Optics) with a narrow depth of field to image the generated interference patterns, and a camera (IDS UI3080) for high-speed acquisition. A tri-band pass filter (Edmund Optics) was mounted in front of the camera to enhance fringe contrast. 

Interferometric recordings of tear films were obtained from human subjects wearing commercial soft contact lenses. Each measurement sequence captured the dynamic evolution of the pre-lens tear film under normal viewing conditions. These dynamic interferograms (Fig.~\ref{fig:InvivoTearFilm_ExperimentalSetup}b) served as the experimental dataset for model training and validation. No explicit control over blinking frequency or tear replenishment was imposed; instead, naturally occurring variations in illumination, eye motion, and fringe contrast were retained to better represent real-world imaging conditions. A representative \textit{in vivo} interferogram and the corresponding manually reconstructed thickness map are shown in Fig.~\ref{fig:InvivoTearFilm_ExperimentalSetup}b. 

\subsection{Dataset Generation}
Due to the limited availability and diversity of experimental interferograms, physiologically relevant synthetic interferograms were developed as follows (Fig.~\ref{fig:InvivoTearFilm_ExperimentalSetup}c). First, an RGB colormap \cite{chandran2020hyperspectral} is generated accounting for the refractive indices of the film, the spectral response of the light source, and the spectral sensitivity of the imaging system (see SI for details). Next, a random thickness profile is generated using either Perlin noise or Gaussian noise. Perlin noise profiles were created with a persistence of 0.5, lacunarity of 1.8, octaves between 1 and 8, and scale between 40 and 150 pixels. Gaussian profiles comprised 30 to 250 superimposed randomly positioned peaks with widths ($\sigma_x$, $\sigma_y$) between 0.1 and 0.5. All profiles were constrained to lie within 0–4000 nm (Fig.~\ref{fig:InvivoTearFilm_ExperimentalSetup}d), with the thickness range inside a profile spanning 250–2500 nm (Fig.~\ref{fig:InvivoTearFilm_ExperimentalSetup}e). Synthetic interferograms were generated by mapping the thickness at each point in the profile to  corresponding color values from the colormap, i.e, at a point $(i,j)$, $\text{Color}(i,j) = \text{Colormap}[\text{thickness}(i,j)]$.

To match experimental conditions, random augmentations are applied to each interferogram including geometric transformations (five-crop, random flips), optical artifacts (pupil masking with diameter 30-50 pixels, shadow effects), Gaussian blur with $\sigma \in [0.1, 3.0]$), color jitter (brightness, contrast, saturation, hue adjustments), and noise injection (Gaussian noise std=10, Poisson noise $\lambda=15$, mean filtering, each at $p=0.5$). The final training dataset comprised a combination of Perlin-based (25-50\%), Gaussian-based (25-50\%), and experimental data (0-50\%). We train on multiple of dataset sizes ranging from 5,000 to 25,000 structures with deterministic random seeds to determine the optimal dataset size at which the model performs best.

\subsection{Model Training}

\begin{figure*}[!h]
\includegraphics[width=\textwidth]{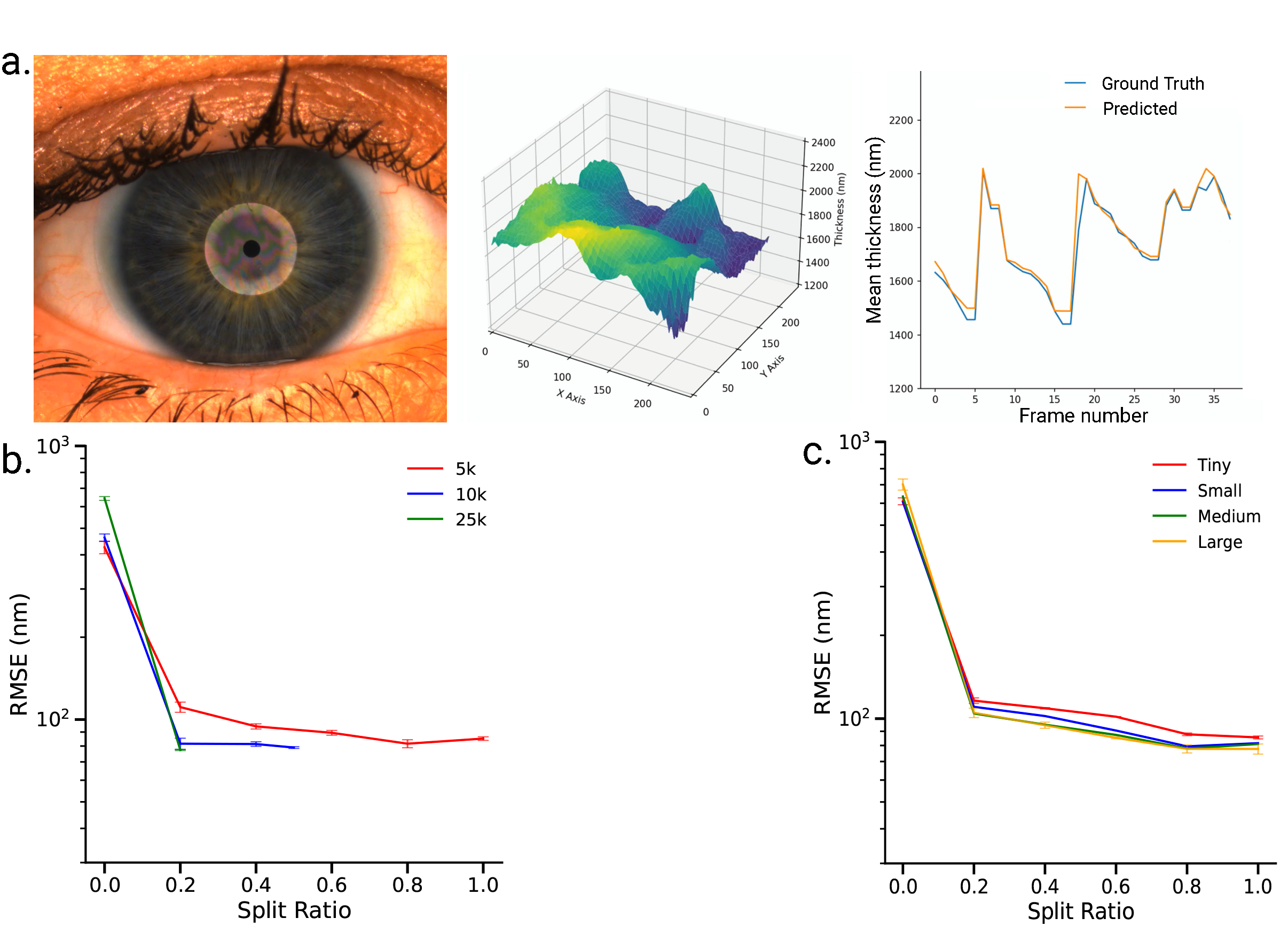}
\caption{Model performance on experimental and synthetic data. (a) Example \textit{in vivo} tear film interferogram, predicted thickness profile, and temporal comparison between predicted and ground truth mean thickness. (b) Effect of dataset size on model accuracy, showing RMSE versus synthetic to experiment data split ratios for datasets totaling 5k, 10k, and 25k samples. (c) Influence of transformer model sizes (Tiny–Large) on prediction accuracy as a function of synthetic to experiment data split ratios for a dataset totaling 5k images.}
\label{fig:ModelPerformance}
\end{figure*}

We train transformer-based depth regressors across a range of learnable model parameters (Tiny, Small, Medium and Large) (Fig.~\ref{fig:ModelPerformance}c). The model parameters were trained to minimize the SiLog loss between the prediction and ground truth. Addition training features included, learn rate optimization with Adam \cite{kingma2014adam} (learning rate $4\times10^{-5}$, weight decay $10^{-2}$, $\epsilon=10^{-3}$), batch size 16, and training over 200 epochs on the datasets described above, with random rotations enabled. Training was performed with distributed data parallelism, and we log training statistics every 100 steps.

\subsection{Model Testing}
Evaluation is conducted on held-out experimental datasets every 500 train steps. Predictions are clamped to the configured evaluation range before scoring (min=0 $\mu$m, max=5 $\mu$m). We report standard depth-estimation metrics: scale-invariant log (SILog), absolute relative error, log10, RMS, squared relative error, log RMS, MAE, MSE, and RMSE. Evaluation is logged in Weights and Biases at fixed intervals (every 500 steps). Final model selection is based on the RMSE computed on the held‑out experimental datasets (lower is better).

\section{Results}\label{sec:Results}
Our best vision transformer-based models effectively reconstructed film thickness across a broad range of thicknesses (up to 3000 nm), successfully resolving multiple interference fringe orders in the presence of imaging noise and ambient artifacts. The model surpasses state-of-the-art in accuracy while achieving an average processing time of ~45 ms per frame on a consumer RTX 3060, enabling near real-time analysis of dynamic thin films. To validate the model's practical performance on human tear film data, we evaluated predictions on experimental interferograms captured \textit{in vivo}. Figure \ref{fig:ModelPerformance}a shows a representative input interferogram, corresponding model prediction, and  a comparison between the predicted and ground truth mean film thickness. The strong correlation between predicted and ground truth thickness profiles demonstrates the model's accuracy in reconstructing tear film topology from real clinical data. Notably, the network successfully handles common experimental artifacts that typically confound traditional analysis methods, including blinking and other involuntary eye movements, partial occlusion by eyelashes at the image periphery, and non-uniform illumination across the curved ocular surface. These results confirm that the vision transformer's learned representations effectively disambiguate phase periodicity even in the presence of realistic imaging imperfections, establishing the method's viability for clinical tear film assessment.

The hybrid training approach combining synthetic and experimental data proved critical for model performance. As shown in Fig.~\ref{fig:ModelPerformance}b, training exclusively on synthetic data resulted in degraded accuracy on experimental images due to unmodeled artifacts like non-uniform illumination and motion blur. Conversely, training solely on limited experimental data restricted generalization across the full thickness range. The combined approach leverages synthetic data's comprehensive coverage while incorporating real-world characteristics from experimental interferograms, yielding robust performance across both domains. Fig.~\ref{fig:ModelPerformance}c demonstrates that model performance has a small impact on tested architecture sizes, with larger models performing marginally better.

\section{Discussion}\label{sec:Conclusion}
We have demonstrated a vision transformer-based approach that successfully addresses the long-standing challenge of real-time thickness reconstruction from thin film interferograms. The key innovation lies in leveraging the transformer's ability to capture long-range spatial dependencies, effectively resolving phase ambiguities that plague conventional phase-unwrapping algorithms.

The hybrid training strategy proved essential for robust performance. Synthetic data provided comprehensive coverage of the thickness space and interference patterns, while experimental data introduced real-world artifacts including motion blur, non-uniform illumination, and blinking related artifacts. This combination enables the model to generalize beyond the limitations of purely synthetic training while avoiding overfitting to the limited experimental dataset. The scalability with model architecture size suggests that further improvements may be achievable with larger transformers, though computational constraints for real-time applications must be considered.

Several limitations warrant future investigation. First, the current model performs poorly on films below 500nm, a critical range where dewetting occurs in tear films. Dewetting represents the final stages of tear film breakup where the aqueous layer fully dries. Dewetted regions are clinically significant as they are correlated to occular discomfort. On going work seeks to address the above limitation. In addition we are also working on quantitatively demonstrating the universality of the model across diverse interferogram sources (eg. bubbles, thin film coatings), and establishing model interpretability.

The clinical implications of this work are significant. Traditional tear film assessment methods like the Schirmer test and fluorescein staining are invasive, provide only qualitative or semi-quantitative information, and cannot capture the dynamic characteristics critical for understanding tear film stability. Our approach enables continuous, non-invasive monitoring with high precision, potentially revolutionizing dry eye disease diagnosis and treatment monitoring. Beyond ophthalmology, this framework is readily adaptable to other thin film applications. Applications in semiconductor manufacturing, coating quality control, and foam stability measurements could benefit from the same approach.

In conclusion, we have presented a machine learning solution that transforms thin film interferometry from a labor-intensive, expert-dependent technique to an automated, real-time diagnostic tool. By combining modern deep learning architectures with physics-informed data generation, we achieve robust performance on challenging experimental data while maintaining the speed necessary for clinical deployment. This work represents a significant step toward making quantitative tear film analysis accessible in clinical practice, with the potential to improve diagnosis and treatment of ocular surface diseases affecting millions worldwide.
\bibliographystyle{vancouver}
\bibliography{References}

\end{document}